\documentclass{svproc}

\usepackage[T1]{fontenc}
\usepackage[utf8]{inputenc}
\usepackage{microtype}
\usepackage{amsmath,amssymb}
\usepackage{booktabs}
\usepackage{multirow}
\usepackage{graphicx}
\usepackage{url}
\usepackage{hyperref}
\hypersetup{colorlinks=true,linkcolor=black,citecolor=black,urlcolor=black}

\usepackage{caption}
\usepackage{booktabs}
\usepackage{multirow}
\usepackage{xcolor}
\usepackage{pifont}
\usepackage{diagbox}
\usepackage{adjustbox}
\usepackage{orcidlink}

\begin{document}
\mainmatter

\title{Non-verbal Real-time Human-AI Interaction in Constrained Robotic Environments}

\titlerunning{Non-Verbal Chat for Constrained HAI}
% If ERF requires anonymization, replace author/institute with placeholders.
\author{Drago\c{s} Costea\thanks{Equal contribution.}\inst{1}\orcidlink{0000-0003-1411-1433} \and Alina Marcu$^{*}$\inst{1}\orcidlink{0000-0002-8534-3728}\and Cristina Laz\u{a}r\inst{1} \orcidlink{0009-0004-6624-2624}\and Marius Leordeanu\inst{1, 2, 3}\orcidlink{0000-0001-8479-8758}}
\authorrunning{Drago\c{s} Costea \and Alina Marcu \and Cristina Laz\u{a}r and Marius Leordeanu}
\institute{
National University of Science and Technology Politehnica of Bucharest, 313 Splaiul Independentei, Bucharest, Romania\\
\email{\{dragos.costea, alina.marcu, cristina.lazar, marius.leordeanu\}@upb.ro}
\and
Institute of Mathematics of the Romanian Academy, Bucharest, Romania 
\and
NORCE - Norwegian Research Center, Norway
\\
%\email{second.author@affil2.org}
}

\maketitle

%%%%%%%%%%%%%%%%%%%%%%%%%%%%%%%%%%%%%%%%%%%%%%%%%%%%%%%%%%%%%%%%%%%%%%%%%%%%%%%%%%%%%%%%%%%%%%%%%%%%%%%%%%%%%%%%%%%%%%%%%%%%%%%%%%%%%%%%%%%%%%%%%%%%%%%%%%%%%%%%%%%%%%%%%%%%%%%%%%%%%%%%%%%%%%%%%%%%%%%%%%%%

\begin{abstract}
We study the ongoing debate regarding the statistical fidelity of AI-generated data compared to human-generated data in the context of non-verbal communication using full body motion. Concretely, we ask if contemporary generative models move beyond surface mimicry to participate in the silent, but expressive dialogue of body language. We tackle this question by introducing the first framework that generates a natural non-verbal interaction between Human and AI in real-time from 2D body keypoints. Our experiments utilize four lightweight architectures which run at up to 100 FPS on an NVIDIA Orin Nano, effectively closing the perception-action loop needed for natural Human-AI interaction. We trained on 437 human video clips and demonstrated that pretraining on synthetically-generated sequences reduces motion errors significantly, without sacrificing speed. Yet, a measurable reality gap persists. When the best model is evaluated on keypoints extracted from cutting-edge text-to-video systems, such as SORA and VEO, we observe that performance drops on SORA-generated clips. However, it degrades far less on VEO, suggesting that temporal coherence, not image fidelity, drives real-world performance. Our results demonstrate that statistically distinguishable differences persist between Human and AI motion. 

\end{abstract}

%%%%%%%%%%%%%%%%%%%%%%%%%%%%%%%%%%%%%%%%%%%%%%%%%%%%%%%%%%%%%%%%%%%%%%%%%%%%%%%%%%%%%%%%%%%%%%%%%%%%%%%%%%%%%%%%%%%%%%%%%%%%%%%%%%%%%%%%%%%%%%%%%%%%%%%%%%%%%%%%%%%%%%%%%%%%%%%%%%%%%%%%%%%%%%%%%%%%%%%%%%%%

\section{Introduction}
\label{sec:intro}

Non-verbal communication, using posture, gesture, and whole-body movement, often conveys affect and intent more effectively than speech~\cite{barrett2016works}. Enabling machines to read and \emph{respond} to these cues is especially important in embodied settings that operate under tight compute, energy, connectivity, and privacy constraints. In such conditions, anticipating \emph{near-future} motion (hundreds of milliseconds to $\sim$1\,s) enables proactive collision avoidance, better handover timing, adaptive camera framing, and more natural joint action.

Despite rapid progress in generative AI, many systems remain “disembodied”: they produce media from prompts but do not engage in continuous, reactive, physically expressive interaction. Existing animation pipelines are largely anchored to speech, such as co-speech gesture generation or audio-driven talking heads~\cite{li2021audio2gestures,zhang2023sadtalker}, leaving purely non-verbal interaction comparatively underexplored, in part due to limited data and the difficulty of modeling subtle person-specific expressions~\cite{zhao2019affective}. 

\vspace{1mm}
\noindent\textbf{Background and relevant work.} In human–robot interaction (HRI), non-verbal behaviors strongly shape first impressions and perceived social qualities, motivating design frameworks that explicitly encode such cues~\cite{urakami2023nonverbal}. On the modeling side, \emph{motion generation} has evolved from recurrent models for short-horizon pose prediction~\cite{martinez2017human} to diffusion models that produce high-fidelity, conditioned motion~\cite{tevet2022human}. In parallel, \emph{emotion recognition} from skeletal data has benefited from spatio-temporal architectures such as ST-GCNs that operate directly on skeleton graphs~\cite{yan2018spatial}. However, generation and recognition are typically treated in isolation. Moreover, widely used affect datasets (e.g. MELD, IEMOCAP) rely on verbal context~\cite{poria2018meld,busso2008iemocap}, while large-scale skeleton/action datasets lack fine-grained emotion labels~\cite{shahroudy2016ntu}. Recent synthetic motion generators (such as MotionLCM~\cite{dai2024motionlcm}) offer scalable pretraining opportunities, and the advent of high-end video generators (like OpenAI's SORA~\cite{sora2024} and Google's VEO3~\cite{veo2024}) raises the question of how well motion understanding models transfer to purely synthetic scenes.

\vspace{1mm}
\noindent\textbf{Contributions.} We close the perception–action loop for non-verbal human–AI interaction by learning from body keypoints, a modality that is both structured and computationally efficient for on-device use. We explore lightweight deep models that jointly (i) \emph{recognize} the affect expressed by an observed human and (ii) \emph{express} affect by predicting low-latency future motion, to enable responsive and transparent behavior without reliance on speech or text. For natural and effective Human-AI interaction, an AI system must not only comprehend human emotion through recognition but also respond appropriately and expressively through generation to favor and engage interaction, to become more welcoming, instead of uncanny. Towards these efforts, we make the following contributions:
\vspace{-1mm}
\begin{itemize}
    \item A dual-task framework for simultaneous, real-time \emph{expression} (future-motion prediction) and \emph{recognition} of non-verbal emotion from body keypoints.
    \item A comparative study of four lightweight architectures tailored to low-latency interaction, establishing robust baselines on authentic human interaction data.
    \item An investigation of large-scale synthetic motion pretraining to boost downstream performance.
    \item An evaluation of trained models on state-of-the-art \emph{synthetic video} (SORA, VEO3) to assess generalization beyond natural footage~\cite{sora2024,veo2024}.
\end{itemize}

%%%%%%%%%%%%%%%%%%%%%%%%%%%%%%%%%%%%%%%%%%%%%%%%%%%%%%%%%%%%%%%%%%%%%%%%%%%%%%%%%%%%%%%%%%%%%%%%%%%%%%%%%%%%%%%%%%%%%%%%%%%%%%%%%%%%%%%%%%%%%%%%%%%%%%%%%%%%%%%%%%%%%%%%%%%%%%%%%%%%%%%%%%%%%%%%%%%%%%%%%%%%

\section{Methodology}
\label{sec:methodology}

This section details our technical approach, from the formal problem definition to the specific deep learning architectures used for full-body motion estimation and emotion recognition, as well as the datasets used for training, pretraining and evaluation. Our goal is to predict future body motion keypoints in real time for a natural human-machine interface. We also classify the underlying emotion from an observed sequence of body keypoints in order for future algorithms to tailor the experience according to subtle motion cues.

\subsection{Problem Formulation}

We address the dual task of non-verbal emotion expression and recognition. \\

\vspace{1mm}
\noindent\textbf{Input:} Our models use a sequence of 2D body keypoints, $X = \{\mathbf{p}_1, \dots, \mathbf{p}_{T_{in}}\}$, representing $T_{in}=60$ frames of motion (equivalent to 2 seconds at 30 FPS). Each frame $\mathbf{p}_t \in \mathbb{R}^{K \times D}$ consists of $K=17$ keypoints in $D=2$ dimensions (x, y coordinates). The 17 keypoints correspond to the standard COCO layout~\cite{lin2014microsoft}, including major joints of the head, torso, arms, and legs.

\noindent\textbf{Output:} The model produces a tuple $(\hat{Y}_{gen}, \hat{y}_{cls})$ containing:
\begin{itemize}
    % \textbf{Body Motion Expression ($\hat{Y}_{gen}$):} 
    \item A predicted sequence of future keypoints, $\hat{Y}_{gen} = \{\hat{\mathbf{p}}_{T_{in}+1}, \dots, \hat{\mathbf{p}}_{T_{in}+T_{out}}\}$, for an output duration of $T_{out}=30$ frames (1 second).
    \item A probability distribution over $C=3$ discrete emotion classes: ``enthusiastic,'' ``laughing,'' and ``happy to see you''.
    % \textbf{Emotion Recognition ($\hat{y}_{cls}$):} 
\end{itemize}

\noindent\textbf{Data preprocessing:} To ensure our models learn motion patterns that are invariant to a person's position and scale in the video frame, we apply a two-step normalization process to each sequence:
\begin{itemize}
\item \textbf{Centering:} We first compute the centroid (mean position) of the 17 keypoints for each frame and subtract it, effectively centering the skeleton at the origin.
\begin{equation}
    \mathbf{p}'_{t,k} = \mathbf{p}_{t,k} - \frac{1}{K}\sum_{j=1}^{K}\mathbf{p}_{t,j}
\end{equation}
\item \textbf{Scaling:} We then normalize the scale by dividing the centered keypoints by the maximum Euclidean distance of any keypoint from the origin within that frame. This maps the pose to a canonical scale.
\begin{equation}
    \mathbf{p}''_{t,k} = \frac{\mathbf{p}'_{t,k}}{\max_{j} ||\mathbf{p}'_{t,j}||_2}
\end{equation}
\end{itemize}

The centroids and scaling factors are stored and used to denormalize the predicted motions for visualization.

\noindent\textbf{Loss function:} Our models are trained end-to-end using a composite loss function that combines a regression loss for motion generation and a classification loss for emotion recognition:
\begin{equation}
    \mathcal{L}_{\text{total}} = \mathcal{L}_{\text{gen}} + \lambda \mathcal{L}_{\text{cls}}
\end{equation}
where $\mathcal{L}_{\text{gen}}$ is the Mean Squared Error (L2 loss) between the predicted and ground truth future keypoints, and $\mathcal{L}_{\text{cls}}$ is the Cross-Entropy loss for emotion classification. After empirical tuning, the classification loss weight $\lambda$ is set to 0.3.

% ----------------------------------------------------------------------------------------------------------------------------

\subsection{Model Architectures}

We benchmark four lightweight architectures, each designed with a shared encoder to create a latent representation of the input motion, followed by two separate heads for the motion generation and emotion recognition tasks.

\begin{itemize}
    \item \textbf{MLP:} A simple Multi-Layer Perceptron baseline. The input sequence of 60 frames is flattened into a single vector of size $60 \times 17 \times 2 = 2040$. This vector is processed by a shared backbone of four fully-connected layers (with 1024, 512, 256, and 128 neurons respectively), each followed by a ReLU activation, Dropout ($p=0.3$), and BatchNorm. The final 128-dimensional feature vector is fed into two heads: a linear layer predicting the flattened 30-frame output sequence, and a 2-layer classifier.

    \item \textbf{LSTM:} A recurrent architecture using a 2-layer LSTM with a hidden size of 128 and a dropout rate of 0.2. Each frame's keypoints are flattened to a 34-dimensional vector, forming an input sequence of length 60. The final hidden state of the LSTM serves as the motion representation, which is then passed to the two task-specific heads.

    \item \textbf{CNN-LSTM:} A hybrid model that first extracts spatio-temporal features using a 1D Convolutional Neural Network. The input sequence (transposed to shape [batch, 34, 60]) is passed through three 1D convolutional layers (with 64, 128, and 256 output channels and a kernel size of 3), each followed by ReLU, BatchNorm, and Dropout. The resulting feature sequence is then fed into a 2-layer LSTM with a hidden size of 128. The final hidden state is used for prediction and classification.

    \item \textbf{Transformer:} A Transformer model using an encoder architecture. The 34-dimensional per-frame input is projected into a 256-dimensional space ($d_{\text{model}}$). We add sinusoidal positional encoding before feeding the sequence into a 4-layer Transformer Encoder with 8 attention heads. The output from the last time step is used as the final feature vector for the two heads.
\end{itemize}

\iffalse
\begin{table}[t]
\centering
\caption{Model inference time comparison on consumer (RTX 4090) and embedded (Orin Nano) GPUs.}
\setlength{\tabcolsep}{12pt}
\begin{tabular}{lccc}
\hline
\textbf{Model} & \textbf{Parameters} & \multicolumn{2}{c}{\textbf{Inference Time [ms]}} \\
 & & \textbf{RTX 4090} & \textbf{Orin Nano} \\
\hline
CNN-LSTM    & 592,060    & 0.832   & 4.844 \\
LSTM        & 347,644    & 0.422   & 3.371 \\
MLP         & 2,914,428  & 0.473   & 1.643 \\
Transformer & 3,430,140  & 1.762   & 9.294 \\
\hline
\end{tabular}
\label{tab:models_latency}
\end{table}
\fi 

All models are small enough to run above real-time frame rates on edge devices, as shown in Table~\ref{tab:models_latency}, the slowest achieving $\approx$100FPS on the NVIDIA Orin Nano. Nevertheless, prediction pipelines could average output sequences for smoother prediction and include keypoint detection, resulting in a longer inference time. 

\begin{figure}[t]
\centering
% Left: the figure (half width)
\begin{minipage}[t]{0.48\linewidth}
  \vspace{0pt} % ensures top baseline alignment
  \centering
  \captionof{figure}{Results confirming the effectiveness of pretraining on AI generated data using MotionLCM~\cite{dai2024motionlcm}. While the choice of architecture is important, most models show significant improvements over the baseline that generally increases with the number of synthetic samples seen - 9k, 45k or 90k, from left to right.}
  \includegraphics[width=\linewidth]{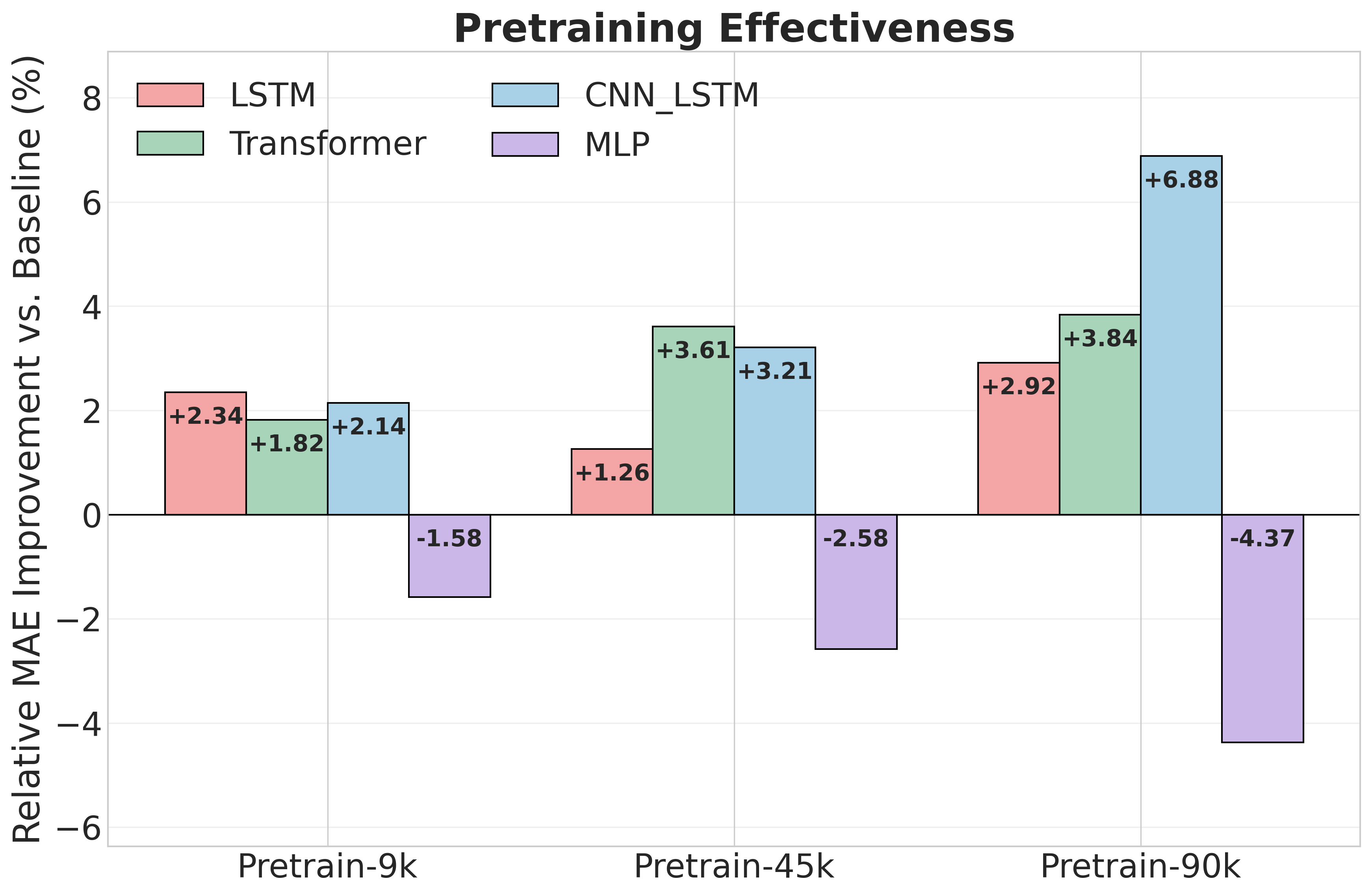}
  \label{fig:pretraining_effectiveness}
\end{minipage}
\hfill
% Right: the table (half width, NON-FLOATING)
\begin{minipage}[t]{0.48\linewidth}
  \vspace{0pt}
  \centering
  \captionof{table}{Model inference time comparison on consumer (\mbox{RTX 4090}) and embedded (\mbox{Orin Nano}) GPUs. While the results show 1-2ms/inference on a desktop GPU, even our largest model runs at about 100 frames per second on an entry level embedded platform, leaving headroom for other operations such as keypoint detection and avatar animation.}
  \label{tab:models_latency}
  % Tighten/loosen as needed for half-width tables:
  \setlength{\tabcolsep}{9pt}       % was 12pt; smaller helps half-width fit
  \renewcommand{\arraystretch}{1.2} % improves readability
  \resizebox{\linewidth}{!}{%
    \begin{tabular}{lccc}
      \hline
      \textbf{Model} & \textbf{Parameters} & \multicolumn{2}{c}{\textbf{Inference Time [ms]}} \\
       & & \textbf{RTX 4090} & \textbf{Orin Nano} \\
      \hline
      CNN-LSTM    & 592,060    & 0.832 & 4.844 \\
      LSTM        & 347,644    & 0.422 & 3.371 \\
      MLP         & 2,914,428  & 0.473 & 1.643 \\
      Transformer & 3,430,140  & 1.762 & 9.294 \\
      \hline
    \end{tabular}%
  }
\end{minipage}
\vspace{-2em}
\end{figure}
% ----------------------------------------------------------------------------------------------------------------------------

\subsection{Datasets}

Our experiments leverage three distinct types of data to train, pretrain and evaluate our models.

\vspace{1mm}
\noindent\textbf{Real Human Motion Dataset:} Our primary dataset consists of 437 video sequences of real human motion depicting highly-positive emotions, annotated with one of the three emotion labels (``enthusiastic,'' ``laughing,'' and ``happy to see you''). 17 keypoints were selected from the 543 generated from MediaPipe's Holistic model~\cite{lugaresi2019mediapipe}, in order to match the COCO format\cite{lin2014microsoft}, for future fast detection. To augment this dataset and create sufficient training samples, we employ a sliding window approach with a 1 frame step. This results in a 16x larger dataset of motion segments (6992 samples), which is then split into an 80\% training set (Train) and a 20\% test set (Test). Throughout our tests we refer to this dataset as either ``real'', ``baseline'' or ``human-generated''. We use these data for both training and finetuning.  

\vspace{1mm}
\noindent\textbf{Synthetic Pretraining Datasets:} To explore the benefit of large-scale data, we generate a synthetic motion dataset using MotionLCM~\cite{dai2024motionlcm}, a text-to-motion diffusion model. MotionLCM only outputs 22-keypoints person skeletons and the data generation process involved prompts only specifying the emotion: "enthusiastic", "laughing" and "happy to see you".  We convert to the COCO keypoints format and generate a 90k sample dataset (30k per emotion) which has been portioned into three progressively larger tiers for pretraining (9k, 45k, 90k samples) termed as Pretrain-9k, Pretrain-45k, and Pretrain-90k. This data is exclusively used in pretraining the models.

\begin{figure}[htbp]
    \centering
    \includegraphics[width=\columnwidth]{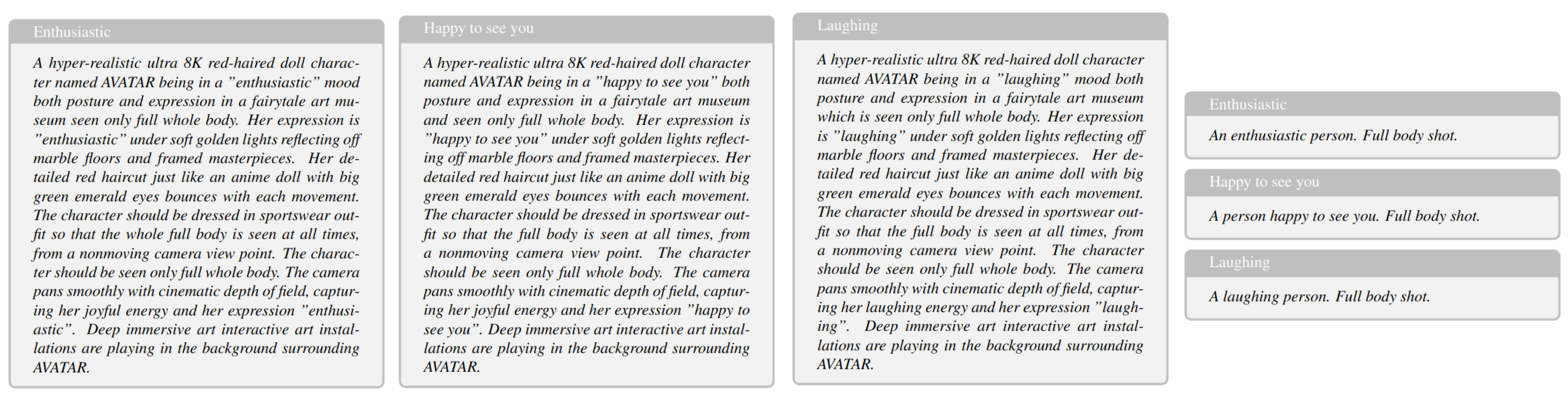}
    \caption{The prompts used to condition SORA (first 3 columns) and VEO (last column) to generation videos depicting each of our proposed emotions.}
    \label{fig:synthetic_data_prompts}
\end{figure}

\vspace{1mm}
\noindent\textbf{Synthetic In-the-Wild Test Datasets (SORA \& VEO):} To test model generalization to unseen, purely synthetic motion, we curate three distinct test sets. These are generated using state-of-the-art text-to-video models: OpenAI's SORA~\cite{sora2024} and Google's VEO3~\cite{veo2024}. We create videos from various prompts (see Figure~\ref{fig:synthetic_data_prompts}) and then apply the same MediaPipe keypoint extraction pipeline used for our real dataset. This provides a challenging ``in-the-wild'' benchmark to measure the gap between models trained on real/MotionLCM data and their ability to interpret motions from different generative sources. The first one, SORA\textsubscript{Human} is generated using the plotted human skeletons from the real data and using them along with a text prompt to generate RGB videos, which are translated back to keypoints using our MediaPipe extraction pipeline described above. The second one, SORA\textsubscript{AI} is generated using the same conditioning procedure, but the input videos are not the real human data skeletons - they are replaced by the ones generated by MotionLCM. Finally, VEO, the third one, is only conditioned on text, consisting on an instruction to generate a human (i.e. "a person"), the emotion being shown (i.e., "enthusiastic", "laughing" and "happy to see you") and camera instructions (i.e. "full body shot"), due to zooming in/ body cut issues we encountered for the emotion-only prompt.

\vspace{-10mm}
%%%%%%%%%%%%%%%%%%%%%%%%%%%%%%%%%%%%%%%%%%%%%%%%%%%%%%%%%%%%%%%%%%%%%%%%%%%%%%%%%%%%%%%%%%%%%%%%%%%%%%%%%%%%%%%%%%%%%%%%%%%%%%%%%%%%%%%%%%%%%%%%%%%%%%%%%%%%%%%%%%%%%%%%%%%%%%%%%%%%%%%%%%%%%
\section{Experiments}
\label{sec:experiments}

We conducted a series of three studies to evaluate our proposed framework. We first establish baseline performance on real data then analyze the effect of synthetic pretraining. Finally, we measure the generalization gap to novel synthetic video content.

\noindent\textbf{Implementation details}: All models were trained using the Adam optimizer for 200 epochs with a batch size of 32. Learning rates were set to 0.001 for LSTM/CNN-LSTM, 0.0005 for Transformer, and 0.002 for MLP, with weight decay of 1e-4. We used PyTorch as our deep-learning framework. All experiments were performed on an Ubuntu 22.04.5 LTS workstation equipped with an Intel Core i9-14900K CPU, 188 GB RAM, and a single NVIDIA RTX 4090 GPU.

\noindent\textbf{Evaluation metrics}: For the body keypoints generation, we report mean absolute error (MAE) for geometric accuracy. For the classification task, we report the mean accuracy. For real-time performance measurements, we report the latency (ms/inference) to validate the real-time capability of our models and also model complexity (in terms of number of parameters) -- see Table~\ref{tab:models_latency}.

\subsection{The effect of pretraining on synthetic data}

In this study, we investigate the hypothesis that pretraining on large-scale synthetic data can improve model performance and data efficiency. 
We first establish the baseline performance by training all four architectures (MLP, LSTM, CNN-LSTM, Transformer) from scratch solely on our real human motion dataset, using the split presented before. The objective is to identify the most effective and efficient architecture for real-time keypoint generation and emotion recognition without leveraging any external synthetic data. 

After this, we employ a pretraining procedure using the same arhitectures, trained from scratch, followed by a finetuning strategy. The model is first pretrained on the synthetic MotionLCM dataset and subsequently finetuned on the real human motion training set.

We systematically evaluate this effect using our three tiers of synthetic data: 9k, 45k, and 90k samples. The results compare the performance of the model fine-tuned from each pretraining tier against the baseline model trained from scratch. Relative performance improvements on the real test sets, compared to the baseline are reported in Figure~\ref{fig:pretraining_effectiveness}. The results indicate that the magnitude of the gain depends on the sequence‑modeling capacity of the backbone architecture and the amount of synthetic pretraining. Recurrent and attention‑based models (LSTM, CNN‑LSTM, Transformer) benefit the most, since their motion prediction error on the real test split consistently drops after pretraining and continues to improve as the pretraining set grows to 90k examples. In contrast, the MLP baseline sees little to no benefit and can even regress, suggesting that purely feed‑forward encoders struggle to absorb the diverse temporal statistics present in MotionLCM outputs. Overall, Figure~\ref{fig:pretraining_effectiveness} supports the view that synthetic motion is most useful when the downstream architecture can explicitly model temporal dependencies and that larger pretraining corpora improve data efficiency for those models.

\subsection{Measuring the synthetic to real gap}

\begin{figure}[htbp]
    \centering
    \includegraphics[width=\columnwidth]{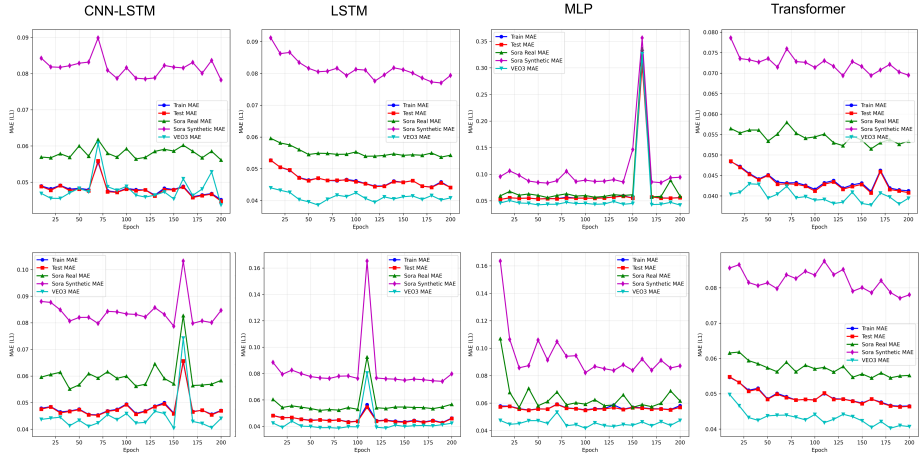}
    \caption{"In-the-wild" evaluation (using MAE) for motion prediction on both synthetic and real data of baseline models (first row) and pretrained models (Pretrain-90k, second row), sampled every 10 epochs. The hierarchy is mostly kept regardless of the architecture or pretraining data. Remarkably, VEO has the lowest error. This suggests that recent large models are more predictable and that specific models might be identified by the way they predict keypoints.}
    \label{fig:pretraining_epochs}
\end{figure}

This study assesses the generalization capabilities of our best model (pretrained on 90k synthetic samples and fine-tuned on real data) by testing it "in-the-wild" on our curated test sets from SORA and VEO. This experiment is designed to quantify the domain gap between the training data distribution (real and MotionLCM) and motions generated by cutting-edge, general-purpose video models. Results are reported in Figure~\ref{fig:pretraining_epochs}.

The figure plots MAE (lower is better) every 10 epochs for each architecture, comparing models trained from scratch (top row) with those pretrained on the 90k synthetic set and then fine‑tuned on real data (bottom row). Across architectures, pretraining shifts the curves downward on the Human test split and accelerates convergence, meaning good performance is reached earlier and maintained more stably, which is desirable for rapid iteration on embedded platforms. When evaluated “in‑the‑wild,” the pretrained models still incur a domain gap, but its size differs by generator: errors on VEO are consistently closer to the real‑data curves than errors on SORA, indicating that temporal coherence and physically plausible dynamics matter more to our keypoint‑based predictor than photorealism. Architecturally, LSTM and CNN‑LSTM exhibit the smoothest and most stable improvements; the Transformer benefits as well but remains more variable; the MLP shows the least consistent gains and occasional spikes, stressing the importance of a model's capacity to learn sequential dependencies.

%%%%%%%%%%%%%%%%%%%%%%%%%%%%%%%%%%%%%%%%%%%%%%%%%%%%%%%%%%%%%%%%%%%%%%%%%%%%%%%%%%%%%%%%%%%%%%%%%%%%%%%%%%%%%%%%%%%%%%%%%%%%%%%%%%%%%%%%%%%%%%%%%%%%%%%%%%%%%%%%%%%%%%%%%%%%%%%%%%%%%%%%%%%%%

\begin{table}[t]
\centering
\caption{Impact of pretraining for the emotion recognition task on various AI-generated test sets (SORA\textsubscript{Human}, SORA\textsubscript{AI} and VEO3) and human-generated test data (denoted with Human in the table). We report Accuracy ($\times 100$). Models are marked by (1) CNN-LSTM, (2) LSTM, (3) MLP, and (4) Transformer.}
\label{tab:classification_results}
\setlength{\tabcolsep}{2pt}      % tighten intercolumn space (default ~6pt)
\renewcommand{\arraystretch}{1.05}
\footnotesize                    % or \scriptsize if needed
\begin{adjustbox}{max width=\textwidth} % auto-shrink to fit line width
\begin{tabular}{@{}l*{4}{cccc}@{}}      % @{} removes outer padding
\toprule
& \multicolumn{4}{c}{SORA\textsubscript{Human}} & \multicolumn{4}{c}{SORA\textsubscript{AI}}
& \multicolumn{4}{c}{VEO} & \multicolumn{4}{c}{Human} \\
\cmidrule(lr){2-5}\cmidrule(lr){6-9}\cmidrule(lr){10-13}\cmidrule(lr){14-17}
& (1) & (2) & (3) & (4) & (1) & (2) & (3) & (4)
& (1) & (2) & (3) & (4) & (1) & (2) & (3) & (4) \\
\midrule
Baseline
& 45.76 & 45.54 & 47.99 & 46.65
& 31.25 & 32.66 & 39.96 & 35.74
& 39.27 & 61.15 & 63.44 & 53.12
& 100   & 100   & 91.71 & 100 \\
Pretrained
& 38.28 & 44.31 & 36.72 & 45.98
& 28.35 & 50.00 & 43.93 & 48.15
& 51.67 & 59.48 & 55.42 & 53.65
& 100   & 100   & 94.78 & 100 \\
$\Delta$ Improvement
& \textcolor{red}{-7.48} & \textcolor{red}{-1.23} & \textcolor{red}{-11.27} & \textcolor{red}{-0.67}
& \textcolor{red}{-2.90} & \textcolor{green!70!black}{+17.34} & \textcolor{green!70!black}{+3.97} & \textcolor{green!70!black}{+12.41}
& \textcolor{green!70!black}{+12.40} & \textcolor{red}{-1.67} & \textcolor{red}{-8.02} & \textcolor{green!70!black}{+0.53}
& - & - & \textcolor{green!70!black}{+3.07} & - \\
\bottomrule
\end{tabular}
\end{adjustbox}
\end{table}

Table~\ref{tab:classification_results} reports classification accuracy for the same four backbones (CNN‑LSTM, LSTM, MLP, Transformer) across SORA\textsubscript{Human}, SORA\textsubscript{AI}, VEO, and Human. 

We observe that on the Human split, accuracy is at or near ceiling for all models, leaving little headroom and hinting at a dataset‑difficulty ceiling effect. Also on SORA\textsubscript{AI}, where SORA is conditioned on MotionLCM skeletons (closer to the pretraining distribution), pretraining substantially helps sequence models, such as the LSTM and Transformer. The CNN‑LSTM is the exception here, possibly reflecting an over‑commitment to pretraining priors that do not perfectly match SORA’s rendering‑to‑keypoint pipeline. 

Our experiments yielded mixed results on SORA\textsubscript{Human} and VEO. On SORA\textsubscript{Human} (SORA conditioned on human skeletons), pretraining can hurt recognition for models, with small negative/neutral changes for the others. This suggests that stylization artifacts introduced when SORA “re‑renders” human motion can push the resulting keypoint dynamics away from both real and MotionLCM statistics. On VEO, the picture is more balanced: CNN‑LSTM improves significantly, the Transformer is essentially unchanged while the LSTM and MLP mildly degrade. Together with Figure~\ref{fig:pretraining_epochs}, these results support our hypothesis that temporal stability and kinematic plausibility (stronger in VEO) better support downstream understanding than image fidelity alone (stronger in SORA). 

The key takeaway is that pretraining on MotionLCM robustly helps generation and selectively helps recognition, with the largest gains when the target motions share generative lineage or temporal statistics with the pretraining source (SORA\textsubscript{AI}, parts of VEO). Conversely, when synthetic renderers perturb human motion in ways foreign to both real and MotionLCM dynamics (SORA\textsubscript{Human}), recognition can degrade, underscoring a measurable—and model‑dependent—synthetic‑to‑real gap.

\textbf{Limitations and Future Work:}
Our study is limited to 2D keypoint prediction and a specific synthetic data generation method for pretraining (MotionLCM). Future work should explore 3D motion prediction and different synthetic data generation approaches. Additionally, the current setup is offline and the evaluation is constrained to only three positive-valence emotions, which may not capture the full diversity of human motion and emotion patterns. A key next step is to integrate this into a live, interactive system with a human-in-the-loop to evaluate user experience and a wider range of emotions. 

\section{Conclusion}
\label{sec:conclusion}
Our study yields three practical insights for non‑verbal HAI. First, synthetic motion is an effective pretraining signal for real‑time models: sequence‑aware backbones (LSTM, CNN‑LSTM, Transformer) reduce motion error on real clips without sacrificing latency, enabling responsive deployment on edge devices. Second, a domain gap persists between human motion and motions rendered by current video generators; it is smaller for VEO than SORA, pointing to the primacy of temporal coherence over image fidelity for keypoint‑level understanding. Third, recognition benefits are architecture and also source dependent: pretraining helps most when the downstream data shares the temporal statistics of the synthetic corpus (such as SORA\textsubscript{AI} and parts of VEO), while stylized re‑renderings of human motion (SORA\textsubscript{Human}) can hurt accuracy. Together with our quantitative analysis, these results suggest a practical recipe: pair large‑scale synthetic pretraining with light real‑world fine‑tuning and favor sequence‑aware models, while developing diagnostics and metrics that explicitly target temporal plausibility to better detect and bridge residual synthetic–real gaps.

\noindent\textbf{Acknowledgements.} This work is supported in part by projects “Romanian Hub for Artificial Intelligence - HRIA”, Smart Growth, Digitization and Financial Instruments Program, 2021-2027 (MySMIS no. 334906) and "European Lighthouse of AI for Sustainability - ELIAS", Horizon Europe program (Grant No. 101120237).

% ------------- Minimal references (inline for simplicity) -------------
{
    \small
    \bibliographystyle{styles/bibtex/splncs_srt}
    \bibliography{refs}
}

\end{document}